\documentclass[11pt]{article}

\usepackage[final]{acl}

\usepackage{times}
\usepackage{latexsym}
\usepackage{amsmath}
\usepackage{booktabs}
\usepackage{array}
\usepackage[T1]{fontenc}

\usepackage[utf8]{inputenc}
\usepackage{amsfonts}
\usepackage{maths}

\usepackage{microtype}

\usepackage{inconsolata}

\usepackage{graphicx}

\title{Signatures of Steerability in Activation Space of Language Models}

\author{
  \textbf{Prajjwal Bhattarai\textsuperscript{1,2}},
  \textbf{Tuka Alhanai\textsuperscript{1}},
\\
  \textsuperscript{1}New York University Abu Dhabi, Abu Dhabi, UAE \\
  \textsuperscript{2}New York University, Tandon School of Engineering, Brooklyn, NY, USA
\\
  \small{
    \textbf{Correspondence:} \href{mailto:pb2276@nyu.edu}{pb2276@nyu.edu}
  }
}

\begin{document}
\maketitle
\begin{abstract}
Steering language models using a set of contrastive representations has been a canonical and computationally efficient method for controlling model behavior. Despite this success in controlling certain model behaviors, the effectiveness of activation steering varies markedly across concepts; the generalization properties of steering vectors are often considered a function of the dataset used to construct them. We make this dataset-dependence claim more rigorous and show that simple separation metrics strongly correlate with the downstream steerability of language models across diverse settings, even after controlling for layers and dataset effects. Beyond prediction, we provide evidence from a synthetic superposition experiment that separation metrics are strongly correlated with alignment between the empirical and true feature direction. Our results suggest that simple separability statistics can serve as practical diagnostics for when steering vectors are likely to work. \footnote{Code available at \url{https://github.com/x-labs-xyz/steering-signatures}}
\end{abstract}

\section{Introduction}

Recent work in language model interpretability shows that model behavior can be altered by adding a linear direction, called a \emph{steering vector}, to hidden activations at inference time. Prior work has shown that such steering vectors can control honesty, harmlessness, and refusal without updating model weights, suggesting that high-level properties of model behavior are encoded, at least approximately, in a one-dimensional subspace of the representation space \citep{subramani2022extracting,turner2023steering,zou2023representation,todd2024function}.

The initial wave of optimism about steering vectors has subsequently been dampened by recent works that have made it clear that steering is not uniformly reliable. Steering effectiveness can vary substantially across concepts, prompt templates, datasets, and intervention layers, and can degrade under reasonable distribution shift \citep{tan2024analysing, silva-etal-2025-steering}. In particular, prior analyses suggest that steerability is often better explained by properties of the source dataset than by the model, raising the possibility that a steering vector succeeds only when the target concept is cleanly expressed in the activations from which it is estimated \citep{tan2024analysing}. 
\begin{figure*}
    \centering
    \includegraphics[width=1\linewidth]{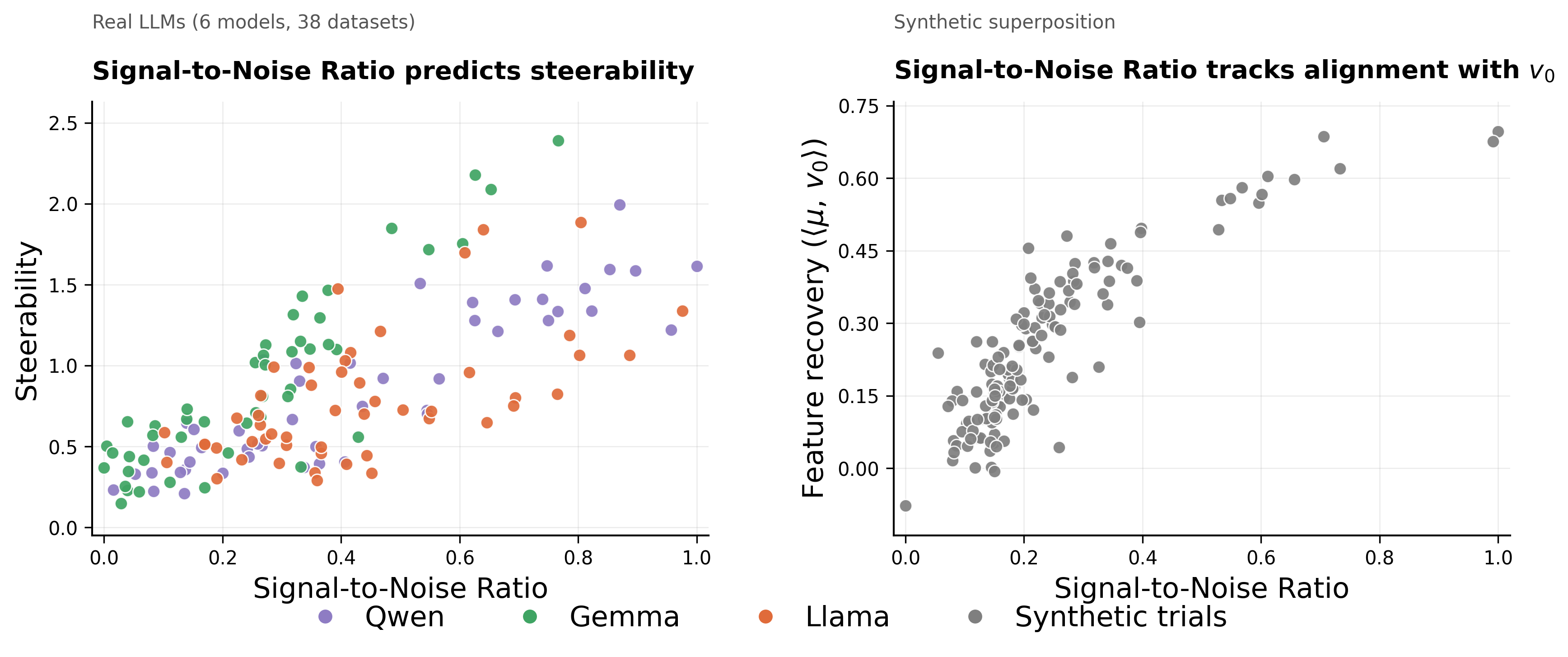}
    \caption{Signal-to-Noise ratio predicts steerability in real models (left) because it detects the recovery of the underlying behavioral direction in controlled superposition (right). Across six instruction-tuned models and 38 persona, truthfulness and safety datasets, SNR explains a substantial fraction of variance in steerability ($r = 0.76$). In a synthetic setting where the ground-truth feature direction $v_0$ is known by construction, SNR is strongly correlated with feature recovery $\pi = \frac{\left\langle \mu,  v_0 \right\rangle}{\norm{\mu}_2}  (\rho = 0.91 )$, but only weakly with sparsity $(\rho = -0.33)$.}
    \label{fig:fig1}
\end{figure*}
These observations motivate the central question of this work: Can we identify a function of the set of activations used to train steering vectors that can predict their downstream performance?  In the context of steering vectors trained through Contrastive Activation Addition (CAA)  \cite{rimsky-etal-2024-steering}, we hypothesize that its downstream reliability should depend on how strongly positive and negative sets of activations are "separated" in the activation space. We make the notion of "separation" rigorous by considering metrics based on (i) directional alignment between steering vector and individual contrastive points, (ii) signal-to-noise ratios that compare the mean difference to class variances, and (iii) geometric or manifold-based measures that capture the structure of activation clouds.

We show that simple separability metrics correlate consistently with steerability across model families, even after controlling for major confounders. In a synthetic superposition setup, we find that the same metrics weakly correlate with the sparsity of the superposition structure, but strongly track the directional alignment between linearly recovered and true feature vector.

\section{Background}
\subsection{Predicting ability from representations}
The relationship between representational geometry and downstream capabilities has been widely studied. Prior work links phenomena such as neural collapse \citep{papyan2020prevalence}, low intrinsic dimension \citep{ansuini2019intrinsic}, and manifold geometry \citep{choudiagnosing} to improved generalization and performance. These geometric diagnostics have largely been studied in image models, with limited applications in language models.

Since steering operates directly in activation space, it is reasonable to expect its success to depend on properties of the underlying representations. \citet{braun2025unreliability} show that the alignment between individual activations and the steering direction strongly correlates with steerability in Llama models. We extend this line of analysis by examining a broader set of separation metrics across multiple model families.
\subsection{Linear Representation Hypothesis \& Superposition}
 Steering is often motivated through the Linear Representation Hypothesis (LRH), the claim that  meaningful concepts are represented as approximately linear directions in the hidden states of language models \citep{park2024lrh}. Empirical work overwhelmingly supports this view; compact directions in activation space have been shown to encode abstractions that are causally relevant for model outputs \citep{bolukbasi2016man, zou2023representation,todd2024function}. However, LRH does not imply that every concept can be cleanly decomposed into unique orthogonal directions. A central challenge is superposition, where models represent many sparse features in overlapping subspaces  \citep{elhage2022toy,arora-etal-2018-linear}. Interference from superposition can become an issue for steering; even if a target concept is linearly present, the direction recovered from finite contrastive data may be contaminated by other latent factors that co-occur in the dataset. We supplement our empirical analysis with a synthetic superposition setting that enables us to explicitly investigate which aspects of superposition our separation metrics capture.
\section{Methods}

\subsection{Dataset \& Models}
We work with 26 multiple choice persona datasets  from \citet{perez2023discovering}, 5 truthfulness datasets from \citet{ying2026truthfulness} and 7 safety datasets from \cite{zhang2024safetybench}. The full list of datasets is included in Appendix \ref{sec:datasets}. Each data point in the dataset contains a triplet $\mathcal{D} = \{x_i, y_i^+, y_i^- \}_{i=1}^{n}$ where $x_i$ is the prompt eliciting a particular persona, $y^+$ is the desired completion and $y^-$ is the undesired completion. For each dataset, we perform a 50-50 train-test split. The train split is used to construct the steering vector and compute activation separation metrics while steerability is computed on the test set.

We experiment with instruction tuned models across two parameter scales (1B-4B and 7B-14B) and across Qwen \cite{qwen3}, Gemma \cite{gemmateam2025gemma3technicalreport}, and Llama \cite{grattafiori2024llama3herdmodels} model families. All experiments were conducted using PyTorch \cite{10.5555/3454287.3455008} and Transformers \cite{wolf-etal-2020-transformers} on a single H100 GPU.

\subsection{Steerability and Separation metrics}

For each dataset, we construct a steering vector, $\mu$, via Contrastive Activation Addition (CAA) \citep{rimsky-etal-2024-steering} as the mean activation difference between contrastive pairs at the last-token position of residual layer. Following \cite{tan2024analysing}, we measure steerability as the slope of the best linear fit between steering multiplier $\lambda$ and the resulting mean propensity, $ m(x) = \text{logit} (y^+) - \text{logit}(y^-)$ on the test set. Intuitively, steerability captures how reliably increasing the steering strength shifts the model toward the target behavior. We provide the full mathematical details of steering vector construction and steerability in Appendix \ref{sec:steeringvec}.

\subsection{Separation criteria}

Let $X_{l}^{\mathcal{D}+}, X_{l}^{\mathcal{D}-} \in \R^{n \times d}$ denote the matrices of positive and negative activations defined above. We drop the indices for notational simplicity. We correlate steerability for each dataset-layer combination with a range of separability metrics.

\subsubsection{Signal-to-noise Separation (SNR)}

We adapt the classic signal-to-noise ratio where we consider the per-dimensional variances between activations of the same class to be the noise. Formally, we define it as:
\begin{equation*}
    SNR = \frac{\norm{\mu}_2^2}{Tr(\Sigma^+) + Tr(\Sigma^-) }
\end{equation*}
where $\Sigma^{+}$ and $\Sigma^{-}$ denote the covariance matrices of $X^{+}$ and $X^{-}$. SNR has been widely adopted in metric learning \cite{boroujeni2018trace, yuan2019signal} but has seen limited use in activation spaces.
\subsubsection{Mean $\mu$ alignment}
Mean $\mu$ alignment measures how well the individual contrastive pairs align with the constructed steering vector.
\begin{equation*}
    \Bar{c_\Delta} = \frac{1}{n} \sum_{i=1}^n \frac{\langle \mu, X^{+}_{i} - X^{-}_i \rangle}{ \norm{\mu}_2 \norm{X_i^+ - X_i^-}}
\end{equation*}
\subsubsection{Manifold capacity}
$\Mc^{+}$, $\Mc^{-}$ are the convex hulls of the sets of activations in $X^+$ and $X^-$. We use the manifold capacity metric \cite{cohen2020separability, chou2024geometry} which is defined as $\alpha = \frac{2}{N_{\text{crit}}}$
where $N_{\text{crit}}$ is the minimum number of dimensions such that random $N_{\text{crit}}$ dimensional projections of points from $\Mc^+$ and $\Mc^-$ are linearly separable with high probability. We compute $N_{\text{crit}}$ using the empirical estimator of \citet{choudiagnosing}.

\subsubsection{Intrinsic dimension}

We estimate the intrinsic dimension of the manifold formed by the difference vectors 
$X^{+} - X^{-}$ using the Two-NN estimator of \citet{facco2017estimating}. 
Concretely, we consider the set of points $\mathcal{M} = \{ x_i^{+} - x_i^{-} \}_{i=1}^{n} \subset \mathbb{R}^{d}$
and treat $\mathcal{M}$ as samples from an embedded manifold in $\mathbb{R}^{d}$.

\subsection{Superposition}
We complement our empirical study h a synthetic setting to study how separability relates to structural properties of superposition. In $\mathbb{R}^d$, we construct  $m$ unit dictionary features $\{v_i\}_{i=0}^m$ with bounded absolute coherence 
$\max_{i \ne j} \abs{\langle v_i, v_j \rangle} \le \epsilon$. We designate $v_0$ as the 
canonical contrastive direction; the remaining features act as noise. Activations take 
the form
\[
        x = \alpha v_0 + \sum_{i=1}^m \beta_i v_i \label{eq:superpos}
\]
where $\alpha \neq 0$ is drawn from distribution $\mathcal{P}$ (for $X^+$) or distribution $\mathcal{Q}$ (for 
$X^-$). Noise coefficients $\beta_i$ are drawn from a normal distribution  and are sparsified so that only a random $k$ sized subset of $\beta_i$ are nonzero. We construct a contrastive dataset and compute $\mu$ using CAA. Full details of this construction are provided in Appendix \ref{sec:superposition_appendix}.

We then study how separation performance varies with three quantities:
\begin{equation*}
    \epsilon = \max_{i \ne j} \langle v_i, v_j \rangle, \; 
    k = |\{i:\beta_i \neq 0\}|, \:
    \pi = \frac{\left\langle \mu,  v_0 \right\rangle}{\norm{\mu}_2}
\end{equation*}
corresponding to feature coherence, the number of active noise features, and the alignment between the 
mean contrastive direction and $v_0$, respectively.
\begin{table*}[t]
\centering
\footnotesize
\setlength{\tabcolsep}{6pt}
\renewcommand{\arraystretch}{1.1}
\begin{tabular}{lcccccc}
\toprule
& \multicolumn{2}{c}{\textbf{Qwen 3}} & \multicolumn{2}{c}{\textbf{Gemma 3}} & \multicolumn{2}{c}{\textbf{Llama}} \\
\cmidrule(lr){2-3} \cmidrule(lr){4-5} \cmidrule(lr){6-7}
\textbf{Metric} & \textbf{14B} & \textbf{4B} & \textbf{12B} & \textbf{4B} & \textbf{3.1 8B} & \textbf{3.2 3B} \\
\midrule
SNR
& \textbf{0.77 / 0.68}
& \textbf{0.81 / 0.74}
& \textbf{0.82 / 0.65}
& 0.81 / 0.79
& \textbf{0.69 / 0.62}
& \textbf{0.75 / 0.68} \\

Mean $\mu$ Alignment
& 0.70 / 0.69
& 0.78 / 0.75
& 0.81 / 0.66
& \textbf{0.82 / 0.81}
& 0.63 / 0.64
& 0.68 / 0.73 \\

Manifold capacity
& 0.56 / 0.63
& 0.62 / 0.66
& 0.76 / 0.54
& 0.71 / 0.66
& 0.55 / 0.58
& 0.47 / 0.58 \\

Intrinsic dim.
& 0.21 / -0.08
& 0.34 / 0.29
& 0.30 / 0.12
& 0.39 / 0.11
& 0.16 / 0.06$^{\dagger}$
& 0.25 / 0.05$^{\dagger}$ \\

\bottomrule
\end{tabular}
\caption{
Correlation between separation metrics and downstream steerability, grouped by model family.
Each cell reports \textbf{Pearson $r$ / Spearman $\rho$}.
All results are statistically significant ($p < 0.05$) unless otherwise marked with $^\dagger$.
}
\label{tab:table1}
\end{table*}
\section{Results}
\subsection{Separability is correlated with steerability}

\begin{table}[t]
\centering
\footnotesize
\setlength{\tabcolsep}{5pt}
\renewcommand{\arraystretch}{1.1}
\begin{tabular}{lcc}
\toprule
\textbf{Metric} & \textbf{Ctrl. dataset} & \textbf{Ctrl. layers} \\
\midrule
SNR               & \textbf{0.74 / 0.56}  & \textbf{0.74 / 0.72} \\
Mean $\mu$ Alignment       & 0.65 / 0.56  & 0.68 / 0.71 \\
Manifold capacity & 0.47 / 0.40  & 0.60 / 0.67 \\
Intrinsic dim.    & 0.33 / 0.20  & 0.07 / -0.05 \\

\bottomrule
\end{tabular}
\caption{
Correlation for each metric aggregated across all models when introducing controls. Each cell reports \textbf{Pearson $r$ / Spearman $\rho$}.
All results are statistically significant
}
\label{tab:avg-pearson-family}
\end{table}
We correlate steerability for each dataset/layer pair for each model with the metrics introduced in Section 3. Table \ref{tab:table1} shows the model-specific correlation for all combinations of datasets and layers, while Appendix \ref{sec:family_corr} shows the results for each family of datasets. Table \ref{tab:avg-pearson-family} shows the partial correlations across models after controlling for layer and datasets. We compute partial correlations through linear residualization on the dataset or layer indicator variables, and compute the Pearson/Spearman correlation of the separation metrics with the residuals. We include details of this construction and the  partial correlations in Appendix \ref{sec:partial_corr}.

We consistently find that SNR is the best predictor of steerability, even after controlling for dataset and layers. SNR has the strongest Pearson and Spearman correlations across the full set of comparisons (Table \ref{tab:table1}) and after controlling for dataset and layer effects (Table \ref{tab:avg-pearson-family}). Mean $\mu$ Alignment is competitive and generally moves alongside SNR.

Manifold capacity is also broadly informative, and moves in the same direction as SNR and mean $\mu$ alignment. However, it rarely surpasses SNR and mean $\mu$ alignment. While broader manifold-level geometry matters, it is likely that simpler activation statistics capture the practically relevant variation more directly.

Intrinsic dimension seems to be weak in all settings, indicating that the  dimensionality of the manifold of the difference of activations might not be a consistent predictor of downstream steerability. Intrinsic dimension also shows an interesting dataset specific variability that is not present in the other metrics; while intrinsic dimension shows positive correlation with steerability for truthfulness (Table \ref{tab:truth_details}) and safety (Table \ref{tab:safety_details}) datasets, the sign flips and it is negative for persona datasets (Table \ref{tab:persona_details}).
 Its correlation also drops remarkably when controlling for layers (Table~\ref{tab:metrics-controlled-layers}) while showing a stronger signal after controlling for datasets (Table~\ref{tab:metrics-controlled-dataset}). It is likely that the correlation of intrinsic dimension with downstream steerability is tracking a layer-dependent representational geometry, rather than a relationship that generalizes across datasets.

\subsection{Why separation predicts steerability?}

An open question remains: which property of the activation distribution do the separation metrics track, such that they predict downstream steerability. We use the synthetic superposition setup and the assumption of the linear representation hypothesis to ask what property of the activation distribution the separation metrics are tracking. We consider three candidate mechanisms. Under the feature-recovery hypothesis, metrics track $\pi$, the cosine similarity between $\mu$ and the true feature direction, and SNR predicts steerability because steering works only to the extent that $\mu$ approximates the true behavioral direction. Under the sparsity hypothesis, metrics track $k$, the number of active features in superposition, and sparser representations are easier to steer. Under the coherence hypothesis, metrics track $\epsilon$, the maximum alignment between true feature directions; more orthogonal feature dictionaries should support cleaner interventions.

Table \ref{tab:superposition-spearman} shows that SNR and mean alignment strongly correlate with $\pi$, have a modest negative correlation with $k$ and show essentially no connection with $\epsilon$. This lends credence to feature recovery over the other two mechanisms. The metrics that best predict steerability in real LLMs track feature recovery rather than global structural properties of superposition in the synthetic setting. Intrinsic dimension shows the opposite pattern and is the only metric showing an association with $k$ and $\epsilon$. However, the fact that intrinsic dimension is the weakest metric in real LLMs suggests that global superposition geometry is neither necessary nor sufficient for predicting steerability.

\begin{table}[t]
\centering
\footnotesize
\setlength{\tabcolsep}{6pt}
\renewcommand{\arraystretch}{1.1}
\begin{tabular}{lccc}
\toprule
\textbf{Metric} & $\epsilon$ & $k$ & $\pi$ \\
\midrule
SNR                 & -0.01$^\dagger$ & -0.33 & \textbf{0.91} \\
Mean $\mu$ Alignment         & -0.01$^\dagger$ & -0.33 & 0.90 \\
Manifold capacity       & 0.01$^\dagger$  & -0.24 & 0.43 \\
Intrinsic dim. & \textbf{-0.12}  & \textbf{0.37} & -0.42 \\
\bottomrule
\end{tabular}
\caption{
Spearman correlation ($\rho$) between separation metrics and structural properties of the synthetic superposition dictionary.
Entries marked with $^\dagger$ are not statistically significant.
}
\label{tab:superposition-spearman}
\end{table}

\section{Conclusion}
We investigate properties of the activation distributions and show that across multiple model families and datasets, simple separation statistics are strong predictors of steerability. Our synthetic superposition experiments further clarify the role of these metrics; the separation statistics only weakly capture global structural properties of superposition but they strongly track whether the recovered direction aligns with the underlying feature direction.

\section{Limitations}

\subsection{Steering method}
We limit our analysis to CAA, one approach within a broad literature on steering vector construction. It is unclear whether the analyses presented here extend to steering methods that do not rely on the same linearity assumptions used in CAA, such as optimization-based steering or methods based on non-linear interactions.

\subsection{Analyses are correlative, not causal}
Our key results are based on correlations between separation metrics and downstream steerability. However, we do not extend the analysis to determine whether separation causally contributes to improved steerability. It remains unclear whether these separations reflect mechanisms the model actually exploits during steering, or if they are merely geometric properties of the representation space without direct computational relevance. It is possible that both the feature recovery mechanism we propose and steerability are caused by an unknown third variable, which drives the positive correlation we report on.

\subsection{Testing on specified task format}

We limit ourselves to multiple-choice datasets and evaluate steering performance based on changes in option logits. Previous work has shown that this multiple-choice propensity is generally correlated with the expression of the concept in open ended generation \cite{rimsky-etal-2024-steering}, however, steering has also been shown to degrade the fluency of the model on unrelated tasks \cite{pres2024towards}. We do not explicitly test the change in fluency of the model in free-form generation. It remains possible that stronger separation improves steerability while degrading fluency or performance on other tasks.

\section{Ethical Considerations}

Our results show that geometric properties of representations can predict if steering interventions are likely to succeed. This can be used for understanding, controlling, and mitigating unwanted behaviors. However, it can also lower the barrier for adversarially manipulating language models in systematic ways. Identifying easily steerable representations can enable malicious actors to systematically induce biased and harmful behavior.

\section{AI use statement}
We utilized AI assistants for literature search, polishing the language of the paper and coding. All AI assisted code was manually validated by the authors before implementation.

\section{Acknowledgments}
Work supported by CAIR and CQTS funded by Tamkeen
NYUAD RI Award CG010 and CG008, respectively. This research was carried out on the High Performance Computing resources at New York University Abu Dhabi.

\bibliography{custom}

@inproceedings{subramani2022extracting,
  title     = {Extracting Latent Steering Vectors from Pretrained Language Models},
  author    = {Subramani, Nishant and Suresh, Nivedita and Peters, Matthew E.},
  booktitle = {Findings of the Association for Computational Linguistics: ACL 2022},
  pages     = {566--581},
  year      = {2022},
  publisher = {Association for Computational Linguistics},
  doi       = {10.18653/v1/2022.findings-acl.48},
  url       = {https://aclanthology.org/2022.findings-acl.48}
}

@misc{turner2023steering,
  title        = {Steering Language Models With Activation Engineering},
  author       = {Turner, Alexander Matt and Thiergart, Lisa and Leech, Gavin and Udell, David and Vazquez, Juan J. and Mini, Ulisse and MacDiarmid, Monte},
  year         = {2023},
  eprint       = {2308.10248},
  archivePrefix= {arXiv},
  primaryClass = {cs.CL},
  doi          = {10.48550/arXiv.2308.10248},
  url          = {https://arxiv.org/abs/2308.10248}
}

@misc{zou2023representation,
  title        = {Representation Engineering: A Top-Down Approach to AI Transparency},
  author       = {Zou, Andy and Phan, Long and Chen, Sarah and Campbell, James and Guo, Phillip and Ren, Richard and Pan, Alexander and Yin, Xuwang and Mazeika, Mantas and Dombrowski, Ann-Kathrin and Goel, Shashwat and Li, Nathaniel and Byun, Michael J. and Wang, Zifan and Mallen, Alex and Basart, Steven and Koyejo, Sanmi and Song, Dawn and Fredrikson, Matt and Kolter, J. Zico and Hendrycks, Dan},
  year         = {2023},
  eprint       = {2310.01405},
  archivePrefix= {arXiv},
  primaryClass = {cs.LG},
  doi          = {10.48550/arXiv.2310.01405},
  url          = {https://arxiv.org/abs/2310.01405}
}

@inproceedings{park2024lrh,
  title     = {The Linear Representation Hypothesis and the Geometry of Large Language Models},
  author    = {Park, Kiho and Choe, Yo Joong and Veitch, Victor},
  booktitle = {Proceedings of the 41st International Conference on Machine Learning},
  year      = {2024},
  url       = {https://arxiv.org/abs/2311.03658},
  note      = {arXiv:2311.03658}
}

@misc{elhage2022toy,
  title        = {Toy Models of Superposition},
  author       = {Elhage, Nelson and Hume, Tristan and Olsson, Catherine and Schiefer, Nicholas and Henighan, Tom and Kravec, Shauna and Hatfield-Dodds, Zac and Lasenby, Robert and Drain, Dawn and Chen, Carol and Grosse, Roger and McCandlish, Sam and Kaplan, Jared and Amodei, Dario and Wattenberg, Martin and Olah, Christopher},
  year         = {2022},
  eprint       = {2209.10652},
  archivePrefix= {arXiv},
  primaryClass = {cs.LG},
  doi          = {10.48550/arXiv.2209.10652},
  url          = {https://arxiv.org/abs/2209.10652}
}

@inproceedings{todd2024function,
  title     = {Function Vectors in Large Language Models},
  author    = {Todd, Eric and Li, Millicent L. and Sharma, Arnab Sen and Mueller, Aaron and Wallace, Byron C. and Bau, David},
  booktitle = {The Twelfth International Conference on Learning Representations},
  year      = {2024},
  url       = {https://openreview.net/forum?id=AwyxtyMwaG},
  note      = {arXiv:2310.15213}
}

@inproceedings{tan2024analysing,
  title     = {Analysing the Generalisation and Reliability of Steering Vectors},
  author    = {Tan, Daniel Chee Hian and Chanin, David and Lynch, Aengus and Paige, Brooks and Kanoulas, Dimitrios and Garriga-Alonso, Adri{\`a} and Kirk, Robert},
  booktitle = {Advances in Neural Information Processing Systems},
  year      = {2024},
  url       = {https://openreview.net/forum?id=v8X70gTodR}
}

@misc{braun2025unreliability,
  title        = {Understanding (Un)Reliability of Steering Vectors in Language Models},
  author       = {Braun, Joschka and Eickhoff, Carsten and Krueger, David and Bahrainian, Seyed Ali and Krasheninnikov, Dmitrii},
  year         = {2025},
  eprint       = {2505.22637},
  archivePrefix= {arXiv},
  primaryClass = {cs.LG},
  doi          = {10.48550/arXiv.2505.22637},
  url          = {https://arxiv.org/abs/2505.22637}
}

@inproceedings{perez2023discovering,
  title={Discovering language model behaviors with model-written evaluations},
  author={Perez, Ethan and Ringer, Sam and Lukosiute, Kamile and Nguyen, Karina and Chen, Edwin and Heiner, Scott and Pettit, Craig and Olsson, Catherine and Kundu, Sandipan and Kadavath, Saurav and others},
  booktitle={Findings of the association for computational linguistics: ACL 2023},
  pages={13387--13434},
  year={2023}
}

@article{qwen3,
    title={Qwen3 Technical Report}, 
    author={An Yang and Anfeng Li and Baosong Yang and Beichen Zhang and Binyuan Hui and Bo Zheng and Bowen Yu and Chang Gao and Chengen Huang and Chenxu Lv and Chujie Zheng and Dayiheng Liu and Fan Zhou and Fei Huang and Feng Hu and Hao Ge and Haoran Wei and Huan Lin and Jialong Tang and Jian Yang and Jianhong Tu and Jianwei Zhang and Jianxin Yang and Jiaxi Yang and Jing Zhou and Jingren Zhou and Junyang Lin and Kai Dang and Keqin Bao and Kexin Yang and Le Yu and Lianghao Deng and Mei Li and Mingfeng Xue and Mingze Li and Pei Zhang and Peng Wang and Qin Zhu and Rui Men and Ruize Gao and Shixuan Liu and Shuang Luo and Tianhao Li and Tianyi Tang and Wenbiao Yin and Xingzhang Ren and Xinyu Wang and Xinyu Zhang and Xuancheng Ren and Yang Fan and Yang Su and Yichang Zhang and Yinger Zhang and Yu Wan and Yuqiong Liu and Zekun Wang and Zeyu Cui and Zhenru Zhang and Zhipeng Zhou and Zihan Qiu},
    journal = {arXiv preprint arXiv:2505.09388},
    year={2025}
}

@misc{grattafiori2024llama3herdmodels,
      title={The Llama 3 Herd of Models}, 
      author={Aaron Grattafiori and Abhimanyu Dubey and Abhinav Jauhri and Abhinav Pandey and Abhishek Kadian and Ahmad Al-Dahle and Aiesha Letman and Akhil Mathur and Alan Schelten and Alex Vaughan and Amy Yang and Angela Fan and Anirudh Goyal and Anthony Hartshorn and Aobo Yang and Archi Mitra and Archie Sravankumar and Artem Korenev and Arthur Hinsvark and Arun Rao and Aston Zhang and Aurelien Rodriguez and Austen Gregerson and Ava Spataru and Baptiste Roziere and Bethany Biron and Binh Tang and Bobbie Chern and Charlotte Caucheteux and Chaya Nayak and Chloe Bi and Chris Marra and Chris McConnell and Christian Keller and Christophe Touret and Chunyang Wu and Corinne Wong and Cristian Canton Ferrer and Cyrus Nikolaidis and Damien Allonsius and Daniel Song and Danielle Pintz and Danny Livshits and Danny Wyatt and David Esiobu and Dhruv Choudhary and Dhruv Mahajan and Diego Garcia-Olano and Diego Perino and Dieuwke Hupkes and Egor Lakomkin and Ehab AlBadawy and Elina Lobanova and Emily Dinan and Eric Michael Smith and Filip Radenovic and Francisco Guzmán and Frank Zhang and Gabriel Synnaeve and Gabrielle Lee and Georgia Lewis Anderson and Govind Thattai and Graeme Nail and Gregoire Mialon and Guan Pang and Guillem Cucurell and Hailey Nguyen and Hannah Korevaar and Hu Xu and Hugo Touvron and Iliyan Zarov and Imanol Arrieta Ibarra and Isabel Kloumann and Ishan Misra and Ivan Evtimov and Jack Zhang and Jade Copet and Jaewon Lee and Jan Geffert and Jana Vranes and Jason Park and Jay Mahadeokar and Jeet Shah and Jelmer van der Linde and Jennifer Billock and Jenny Hong and Jenya Lee and Jeremy Fu and Jianfeng Chi and Jianyu Huang and Jiawen Liu and Jie Wang and Jiecao Yu and Joanna Bitton and Joe Spisak and Jongsoo Park and Joseph Rocca and Joshua Johnstun and Joshua Saxe and Junteng Jia and Kalyan Vasuden Alwala and Karthik Prasad and Kartikeya Upasani and Kate Plawiak and Ke Li and Kenneth Heafield and Kevin Stone and Khalid El-Arini and Krithika Iyer and Kshitiz Malik and Kuenley Chiu and Kunal Bhalla and Kushal Lakhotia and Lauren Rantala-Yeary and Laurens van der Maaten and Lawrence Chen and Liang Tan and Liz Jenkins and Louis Martin and Lovish Madaan and Lubo Malo and Lukas Blecher and Lukas Landzaat and Luke de Oliveira and Madeline Muzzi and Mahesh Pasupuleti and Mannat Singh and Manohar Paluri and Marcin Kardas and Maria Tsimpoukelli and Mathew Oldham and Mathieu Rita and Maya Pavlova and Melanie Kambadur and Mike Lewis and Min Si and Mitesh Kumar Singh and Mona Hassan and Naman Goyal and Narjes Torabi and Nikolay Bashlykov and Nikolay Bogoychev and Niladri Chatterji and Ning Zhang and Olivier Duchenne and Onur Çelebi and Patrick Alrassy and Pengchuan Zhang and Pengwei Li and Petar Vasic and Peter Weng and Prajjwal Bhargava and Pratik Dubal and Praveen Krishnan and Punit Singh Koura and Puxin Xu and Qing He and Qingxiao Dong and Ragavan Srinivasan and Raj Ganapathy and Ramon Calderer and Ricardo Silveira Cabral and Robert Stojnic and Roberta Raileanu and Rohan Maheswari and Rohit Girdhar and Rohit Patel and Romain Sauvestre and Ronnie Polidoro and Roshan Sumbaly and Ross Taylor and Ruan Silva and Rui Hou and Rui Wang and Saghar Hosseini and Sahana Chennabasappa and Sanjay Singh and Sean Bell and Seohyun Sonia Kim and Sergey Edunov and Shaoliang Nie and Sharan Narang and Sharath Raparthy and Sheng Shen and Shengye Wan and Shruti Bhosale and Shun Zhang and Simon Vandenhende and Soumya Batra and Spencer Whitman and Sten Sootla and Stephane Collot and Suchin Gururangan and Sydney Borodinsky and Tamar Herman and Tara Fowler and Tarek Sheasha and Thomas Georgiou and Thomas Scialom and Tobias Speckbacher and Todor Mihaylov and Tong Xiao and Ujjwal Karn and Vedanuj Goswami and Vibhor Gupta and Vignesh Ramanathan and Viktor Kerkez and Vincent Gonguet and Virginie Do and Vish Vogeti and Vítor Albiero and Vladan Petrovic and Weiwei Chu and Wenhan Xiong and Wenyin Fu and Whitney Meers and Xavier Martinet and Xiaodong Wang and Xiaofang Wang and Xiaoqing Ellen Tan and Xide Xia and Xinfeng Xie and Xuchao Jia and Xuewei Wang and Yaelle Goldschlag and Yashesh Gaur and Yasmine Babaei and Yi Wen and Yiwen Song and Yuchen Zhang and Yue Li and Yuning Mao and Zacharie Delpierre Coudert and Zheng Yan and Zhengxing Chen and Zoe Papakipos and Aaditya Singh and Aayushi Srivastava and Abha Jain and Adam Kelsey and Adam Shajnfeld and Adithya Gangidi and Adolfo Victoria and Ahuva Goldstand and Ajay Menon and Ajay Sharma and Alex Boesenberg and Alexei Baevski and Allie Feinstein and Amanda Kallet and Amit Sangani and Amos Teo and Anam Yunus and Andrei Lupu and Andres Alvarado and Andrew Caples and Andrew Gu and Andrew Ho and Andrew Poulton and Andrew Ryan and Ankit Ramchandani and Annie Dong and Annie Franco and Anuj Goyal and Aparajita Saraf and Arkabandhu Chowdhury and Ashley Gabriel and Ashwin Bharambe and Assaf Eisenman and Azadeh Yazdan and Beau James and Ben Maurer and Benjamin Leonhardi and Bernie Huang and Beth Loyd and Beto De Paola and Bhargavi Paranjape and Bing Liu and Bo Wu and Boyu Ni and Braden Hancock and Bram Wasti and Brandon Spence and Brani Stojkovic and Brian Gamido and Britt Montalvo and Carl Parker and Carly Burton and Catalina Mejia and Ce Liu and Changhan Wang and Changkyu Kim and Chao Zhou and Chester Hu and Ching-Hsiang Chu and Chris Cai and Chris Tindal and Christoph Feichtenhofer and Cynthia Gao and Damon Civin and Dana Beaty and Daniel Kreymer and Daniel Li and David Adkins and David Xu and Davide Testuggine and Delia David and Devi Parikh and Diana Liskovich and Didem Foss and Dingkang Wang and Duc Le and Dustin Holland and Edward Dowling and Eissa Jamil and Elaine Montgomery and Eleonora Presani and Emily Hahn and Emily Wood and Eric-Tuan Le and Erik Brinkman and Esteban Arcaute and Evan Dunbar and Evan Smothers and Fei Sun and Felix Kreuk and Feng Tian and Filippos Kokkinos and Firat Ozgenel and Francesco Caggioni and Frank Kanayet and Frank Seide and Gabriela Medina Florez and Gabriella Schwarz and Gada Badeer and Georgia Swee and Gil Halpern and Grant Herman and Grigory Sizov and Guangyi and Zhang and Guna Lakshminarayanan and Hakan Inan and Hamid Shojanazeri and Han Zou and Hannah Wang and Hanwen Zha and Haroun Habeeb and Harrison Rudolph and Helen Suk and Henry Aspegren and Hunter Goldman and Hongyuan Zhan and Ibrahim Damlaj and Igor Molybog and Igor Tufanov and Ilias Leontiadis and Irina-Elena Veliche and Itai Gat and Jake Weissman and James Geboski and James Kohli and Janice Lam and Japhet Asher and Jean-Baptiste Gaya and Jeff Marcus and Jeff Tang and Jennifer Chan and Jenny Zhen and Jeremy Reizenstein and Jeremy Teboul and Jessica Zhong and Jian Jin and Jingyi Yang and Joe Cummings and Jon Carvill and Jon Shepard and Jonathan McPhie and Jonathan Torres and Josh Ginsburg and Junjie Wang and Kai Wu and Kam Hou U and Karan Saxena and Kartikay Khandelwal and Katayoun Zand and Kathy Matosich and Kaushik Veeraraghavan and Kelly Michelena and Keqian Li and Kiran Jagadeesh and Kun Huang and Kunal Chawla and Kyle Huang and Lailin Chen and Lakshya Garg and Lavender A and Leandro Silva and Lee Bell and Lei Zhang and Liangpeng Guo and Licheng Yu and Liron Moshkovich and Luca Wehrstedt and Madian Khabsa and Manav Avalani and Manish Bhatt and Martynas Mankus and Matan Hasson and Matthew Lennie and Matthias Reso and Maxim Groshev and Maxim Naumov and Maya Lathi and Meghan Keneally and Miao Liu and Michael L. Seltzer and Michal Valko and Michelle Restrepo and Mihir Patel and Mik Vyatskov and Mikayel Samvelyan and Mike Clark and Mike Macey and Mike Wang and Miquel Jubert Hermoso and Mo Metanat and Mohammad Rastegari and Munish Bansal and Nandhini Santhanam and Natascha Parks and Natasha White and Navyata Bawa and Nayan Singhal and Nick Egebo and Nicolas Usunier and Nikhil Mehta and Nikolay Pavlovich Laptev and Ning Dong and Norman Cheng and Oleg Chernoguz and Olivia Hart and Omkar Salpekar and Ozlem Kalinli and Parkin Kent and Parth Parekh and Paul Saab and Pavan Balaji and Pedro Rittner and Philip Bontrager and Pierre Roux and Piotr Dollar and Polina Zvyagina and Prashant Ratanchandani and Pritish Yuvraj and Qian Liang and Rachad Alao and Rachel Rodriguez and Rafi Ayub and Raghotham Murthy and Raghu Nayani and Rahul Mitra and Rangaprabhu Parthasarathy and Raymond Li and Rebekkah Hogan and Robin Battey and Rocky Wang and Russ Howes and Ruty Rinott and Sachin Mehta and Sachin Siby and Sai Jayesh Bondu and Samyak Datta and Sara Chugh and Sara Hunt and Sargun Dhillon and Sasha Sidorov and Satadru Pan and Saurabh Mahajan and Saurabh Verma and Seiji Yamamoto and Sharadh Ramaswamy and Shaun Lindsay and Shaun Lindsay and Sheng Feng and Shenghao Lin and Shengxin Cindy Zha and Shishir Patil and Shiva Shankar and Shuqiang Zhang and Shuqiang Zhang and Sinong Wang and Sneha Agarwal and Soji Sajuyigbe and Soumith Chintala and Stephanie Max and Stephen Chen and Steve Kehoe and Steve Satterfield and Sudarshan Govindaprasad and Sumit Gupta and Summer Deng and Sungmin Cho and Sunny Virk and Suraj Subramanian and Sy Choudhury and Sydney Goldman and Tal Remez and Tamar Glaser and Tamara Best and Thilo Koehler and Thomas Robinson and Tianhe Li and Tianjun Zhang and Tim Matthews and Timothy Chou and Tzook Shaked and Varun Vontimitta and Victoria Ajayi and Victoria Montanez and Vijai Mohan and Vinay Satish Kumar and Vishal Mangla and Vlad Ionescu and Vlad Poenaru and Vlad Tiberiu Mihailescu and Vladimir Ivanov and Wei Li and Wenchen Wang and Wenwen Jiang and Wes Bouaziz and Will Constable and Xiaocheng Tang and Xiaojian Wu and Xiaolan Wang and Xilun Wu and Xinbo Gao and Yaniv Kleinman and Yanjun Chen and Ye Hu and Ye Jia and Ye Qi and Yenda Li and Yilin Zhang and Ying Zhang and Yossi Adi and Youngjin Nam and Yu and Wang and Yu Zhao and Yuchen Hao and Yundi Qian and Yunlu Li and Yuzi He and Zach Rait and Zachary DeVito and Zef Rosnbrick and Zhaoduo Wen and Zhenyu Yang and Zhiwei Zhao and Zhiyu Ma},
      year={2024},
      eprint={2407.21783},
      archivePrefix={arXiv},
      primaryClass={cs.AI},
      url={https://arxiv.org/abs/2407.21783}, 
}

@misc{gemmateam2025gemma3technicalreport,
      title={Gemma 3 Technical Report}, 
      author={Gemma Team and Aishwarya Kamath and Johan Ferret and Shreya Pathak and Nino Vieillard and Ramona Merhej and Sarah Perrin and Tatiana Matejovicova and Alexandre Ramé and Morgane Rivière and Louis Rouillard and Thomas Mesnard and Geoffrey Cideron and Jean-bastien Grill and Sabela Ramos and Edouard Yvinec and Michelle Casbon and Etienne Pot and Ivo Penchev and Gaël Liu and Francesco Visin and Kathleen Kenealy and Lucas Beyer and Xiaohai Zhai and Anton Tsitsulin and Robert Busa-Fekete and Alex Feng and Noveen Sachdeva and Benjamin Coleman and Yi Gao and Basil Mustafa and Iain Barr and Emilio Parisotto and David Tian and Matan Eyal and Colin Cherry and Jan-Thorsten Peter and Danila Sinopalnikov and Surya Bhupatiraju and Rishabh Agarwal and Mehran Kazemi and Dan Malkin and Ravin Kumar and David Vilar and Idan Brusilovsky and Jiaming Luo and Andreas Steiner and Abe Friesen and Abhanshu Sharma and Abheesht Sharma and Adi Mayrav Gilady and Adrian Goedeckemeyer and Alaa Saade and Alex Feng and Alexander Kolesnikov and Alexei Bendebury and Alvin Abdagic and Amit Vadi and András György and André Susano Pinto and Anil Das and Ankur Bapna and Antoine Miech and Antoine Yang and Antonia Paterson and Ashish Shenoy and Ayan Chakrabarti and Bilal Piot and Bo Wu and Bobak Shahriari and Bryce Petrini and Charlie Chen and Charline Le Lan and Christopher A. Choquette-Choo and CJ Carey and Cormac Brick and Daniel Deutsch and Danielle Eisenbud and Dee Cattle and Derek Cheng and Dimitris Paparas and Divyashree Shivakumar Sreepathihalli and Doug Reid and Dustin Tran and Dustin Zelle and Eric Noland and Erwin Huizenga and Eugene Kharitonov and Frederick Liu and Gagik Amirkhanyan and Glenn Cameron and Hadi Hashemi and Hanna Klimczak-Plucińska and Harman Singh and Harsh Mehta and Harshal Tushar Lehri and Hussein Hazimeh and Ian Ballantyne and Idan Szpektor and Ivan Nardini and Jean Pouget-Abadie and Jetha Chan and Joe Stanton and John Wieting and Jonathan Lai and Jordi Orbay and Joseph Fernandez and Josh Newlan and Ju-yeong Ji and Jyotinder Singh and Kat Black and Kathy Yu and Kevin Hui and Kiran Vodrahalli and Klaus Greff and Linhai Qiu and Marcella Valentine and Marina Coelho and Marvin Ritter and Matt Hoffman and Matthew Watson and Mayank Chaturvedi and Michael Moynihan and Min Ma and Nabila Babar and Natasha Noy and Nathan Byrd and Nick Roy and Nikola Momchev and Nilay Chauhan and Noveen Sachdeva and Oskar Bunyan and Pankil Botarda and Paul Caron and Paul Kishan Rubenstein and Phil Culliton and Philipp Schmid and Pier Giuseppe Sessa and Pingmei Xu and Piotr Stanczyk and Pouya Tafti and Rakesh Shivanna and Renjie Wu and Renke Pan and Reza Rokni and Rob Willoughby and Rohith Vallu and Ryan Mullins and Sammy Jerome and Sara Smoot and Sertan Girgin and Shariq Iqbal and Shashir Reddy and Shruti Sheth and Siim Põder and Sijal Bhatnagar and Sindhu Raghuram Panyam and Sivan Eiger and Susan Zhang and Tianqi Liu and Trevor Yacovone and Tyler Liechty and Uday Kalra and Utku Evci and Vedant Misra and Vincent Roseberry and Vlad Feinberg and Vlad Kolesnikov and Woohyun Han and Woosuk Kwon and Xi Chen and Yinlam Chow and Yuvein Zhu and Zichuan Wei and Zoltan Egyed and Victor Cotruta and Minh Giang and Phoebe Kirk and Anand Rao and Kat Black and Nabila Babar and Jessica Lo and Erica Moreira and Luiz Gustavo Martins and Omar Sanseviero and Lucas Gonzalez and Zach Gleicher and Tris Warkentin and Vahab Mirrokni and Evan Senter and Eli Collins and Joelle Barral and Zoubin Ghahramani and Raia Hadsell and Yossi Matias and D. Sculley and Slav Petrov and Noah Fiedel and Noam Shazeer and Oriol Vinyals and Jeff Dean and Demis Hassabis and Koray Kavukcuoglu and Clement Farabet and Elena Buchatskaya and Jean-Baptiste Alayrac and Rohan Anil and Dmitry and Lepikhin and Sebastian Borgeaud and Olivier Bachem and Armand Joulin and Alek Andreev and Cassidy Hardin and Robert Dadashi and Léonard Hussenot},
      year={2025},
      eprint={2503.19786},
      archivePrefix={arXiv},
      primaryClass={cs.CL},
      url={https://arxiv.org/abs/2503.19786}, 
}

@inproceedings{silva-etal-2025-steering,
    title = "Steering off Course: Reliability Challenges in Steering Language Models",
    author = "Da Silva, Patrick Queiroz  and
      Sethuraman, Hari  and
      Rajagopal, Dheeraj  and
      Hajishirzi, Hannaneh  and
      Kumar, Sachin",
    editor = "Che, Wanxiang  and
      Nabende, Joyce  and
      Shutova, Ekaterina  and
      Pilehvar, Mohammad Taher",
    booktitle = "Proceedings of the 63rd Annual Meeting of the Association for Computational Linguistics (Volume 1: Long Papers)",
    month = jul,
    year = "2025",
    address = "Vienna, Austria",
    publisher = "Association for Computational Linguistics",
    url = "https://aclanthology.org/2025.acl-long.974/",
    doi = "10.18653/v1/2025.acl-long.974",
    pages = "19856--19882",
    ISBN = "979-8-89176-251-0"
}

@inproceedings{rimsky-etal-2024-steering,
    title = "Steering Llama 2 via Contrastive Activation Addition",
    author = "Rimsky, Nina  and
      Gabrieli, Nick  and
      Schulz, Julian  and
      Tong, Meg  and
      Hubinger, Evan  and
      Turner, Alexander",
    editor = "Ku, Lun-Wei  and
      Martins, Andre  and
      Srikumar, Vivek",
    booktitle = "Proceedings of the 62nd Annual Meeting of the Association for Computational Linguistics (Volume 1: Long Papers)",
    month = aug,
    year = "2024",
    address = "Bangkok, Thailand",
    publisher = "Association for Computational Linguistics",
    url = "https://aclanthology.org/2024.acl-long.828/",
    doi = "10.18653/v1/2024.acl-long.828",
    pages = "15504--15522"
}

@article{bolukbasi2016man,
  title={Man is to computer programmer as woman is to homemaker? debiasing word embeddings},
  author={Bolukbasi, Tolga and Chang, Kai-Wei and Zou, James Y and Saligrama, Venkatesh and Kalai, Adam T},
  journal={Advances in neural information processing systems},
  volume={29},
  year={2016}
}

@article{arora-etal-2018-linear,
    title = "Linear Algebraic Structure of Word Senses, with Applications to Polysemy",
    author = "Arora, Sanjeev  and
      Li, Yuanzhi  and
      Liang, Yingyu  and
      Ma, Tengyu  and
      Risteski, Andrej",
    editor = "Lee, Lillian  and
      Johnson, Mark  and
      Toutanova, Kristina  and
      Roark, Brian",
    journal = "Transactions of the Association for Computational Linguistics",
    volume = "6",
    year = "2018",
    address = "Cambridge, MA",
    publisher = "MIT Press",
    url = "https://aclanthology.org/Q18-1034/",
    doi = "10.1162/tacl_a_00034",
    pages = "483--495"
}

@inproceedings{boroujeni2018trace,
  title={Trace ratio optimization with feature correlation mining for multiclass discriminant analysis},
  author={Boroujeni, Forough Rezaei and Wang, Sen and Li, Zhihui and West, Nicholas and Stantic, Bela and Yao, Lina and Long, Guodong},
  booktitle={Proceedings of the AAAI Conference on Artificial Intelligence},
  volume={32},
  number={1},
  year={2018}
}

@inproceedings{yuan2019signal,
  title={Signal-to-noise ratio: A robust distance metric for deep metric learning},
  author={Yuan, Tongtong and Deng, Weihong and Tang, Jian and Tang, Yinan and Chen, Binghui},
  booktitle={Proceedings of the IEEE/CVF conference on computer vision and pattern recognition},
  pages={4815--4824},
  year={2019}
}

@article{cohen2020separability,
  title={Separability and geometry of object manifolds in deep neural networks},
  author={Cohen, Uri and Chung, SueYeon and Lee, Daniel D and Sompolinsky, Haim},
  journal={Nature communications},
  volume={11},
  number={1},
  pages={746},
  year={2020},
  publisher={Nature Publishing Group UK London}
}

@article{chou2024geometry,
  title={Geometry linked to untangling efficiency reveals structure and computation in neural populations},
  author={Chou, Chi-Ning and Kim, Royoung and Arend, Luke A and Yang, Yao-Yuan and Mensh, Brett D and Shim, Won Mok and Perich, Matthew G and Chung, SueYeon},
  journal={bioRxiv},
  pages={2024--02},
  year={2024},
  publisher={Cold Spring Harbor Laboratory}
}

@inproceedings{
choudiagnosing,
title={Diagnosing Generalization Failures from Representational Geometry Markers},
author={Chi-Ning Chou and Artem Kirsanov and Yao-Yuan Yang and SueYeon Chung},
booktitle={The Fourteenth International Conference on Learning Representations},
year={2026},
url={https://openreview.net/forum?id=c2fQBcoKhU}
}

@article{facco2017estimating,
  title={Estimating the intrinsic dimension of datasets by a minimal neighborhood information},
  author={Facco, Elena and d’Errico, Maria and Rodriguez, Alex and Laio, Alessandro},
  journal={Scientific reports},
  volume={7},
  number={1},
  pages={12140},
  year={2017},
  publisher={Nature Publishing Group UK London}
}

@article{papyan2020prevalence,
  title={Prevalence of neural collapse during the terminal phase of deep learning training},
  author={Papyan, Vardan and Han, XY and Donoho, David L},
  journal={Proceedings of the National Academy of Sciences},
  volume={117},
  number={40},
  pages={24652--24663},
  year={2020},
  publisher={National Academy of Sciences}
}

@article{ansuini2019intrinsic,
  title={Intrinsic dimension of data representations in deep neural networks},
  author={Ansuini, Alessio and Laio, Alessandro and Macke, Jakob H and Zoccolan, Davide},
  journal={Advances in Neural Information Processing Systems},
  volume={32},
  year={2019}
}

@article{pres2024towards,
  title={Towards reliable evaluation of behavior steering interventions in LLMs},
  author={Pres, Itamar and Ruis, Laura and Lubana, Ekdeep Singh and Krueger, David},
  journal={arXiv preprint arXiv:2410.17245},
  year={2024}
}

@article{ying2026truthfulness,
  title={The Truthfulness Spectrum Hypothesis},
  author={Ying, Zhuofan Josh and Ravfogel, Shauli and Kriegeskorte, Nikolaus and Hase, Peter},
  journal={arXiv preprint arXiv:2602.20273},
  year={2026}
}

@inproceedings{zhang2024safetybench,
  title={Safetybench: Evaluating the safety of large language models},
  author={Zhang, Zhexin and Lei, Leqi and Wu, Lindong and Sun, Rui and Huang, Yongkang and Long, Chong and Liu, Xiao and Lei, Xuanyu and Tang, Jie and Huang, Minlie},
  booktitle={Proceedings of the 62nd Annual Meeting of the Association for Computational Linguistics (Volume 1: Long Papers)},
  pages={15537--15553},
  year={2024}
}

@article{hendrycks2021ethics,
  title={Aligning AI With Shared Human Values},
  author={Dan Hendrycks and Collin Burns and Steven Basart and Andrew Critch and Jerry Li and Dawn Song and Jacob Steinhardt},
  journal={Proceedings of the International Conference on Learning Representations (ICLR)},
  year={2021}
}

@inbook{10.5555/3454287.3455008,
author = {Paszke, Adam and Gross, Sam and Massa, Francisco and Lerer, Adam and Bradbury, James and Chanan, Gregory and Killeen, Trevor and Lin, Zeming and Gimelshein, Natalia and Antiga, Luca and Desmaison, Alban and K\"{o}pf, Andreas and Yang, Edward and DeVito, Zach and Raison, Martin and Tejani, Alykhan and Chilamkurthy, Sasank and Steiner, Benoit and Fang, Lu and Bai, Junjie and Chintala, Soumith},
title = {PyTorch: an imperative style, high-performance deep learning library},
year = {2019},
publisher = {Curran Associates Inc.},
address = {Red Hook, NY, USA},
booktitle = {Proceedings of the 33rd International Conference on Neural Information Processing Systems},
articleno = {721},
numpages = {12}
}

@inproceedings{wolf-etal-2020-transformers,
    title = "Transformers: State-of-the-Art Natural Language Processing",
    author = "Thomas Wolf and Lysandre Debut and Victor Sanh and Julien Chaumond and Clement Delangue and Anthony Moi and Pierric Cistac and Tim Rault and Rémi Louf and Morgan Funtowicz and Joe Davison and Sam Shleifer and Patrick von Platen and Clara Ma and Yacine Jernite and Julien Plu and Canwen Xu and Teven Le Scao and Sylvain Gugger and Mariama Drame and Quentin Lhoest and Alexander M. Rush",
    booktitle = "Proceedings of the 2020 Conference on Empirical Methods in Natural Language Processing: System Demonstrations",
    month = oct,
    year = "2020",
    address = "Online",
    publisher = "Association for Computational Linguistics",
    url = "https://aclanthology.org/2020.emnlp-demos.6/",
    pages = "38--45"
}

\appendix

\section{Mathematical details on steerability and separation metrics}\label{sec:steeringvec}

For a dataset $\mathcal{D}$, we construct a steering vector $\mu_l^{\mathcal{D}}$ by extracting the activations from the last token after residual layer $l$. We construct two contrastive prompts; one with the positive completion $a_l (x,y^+) \in \R^d$ and one with the negative completion $a_l(x,y^-) \in \R^d$. The steering vector is constructed by taking the mean difference between the contrastive pairs.
\begin{equation}
    \mu_l^{\mathcal{D}} = \frac{1}{n} \sum_{(x,y^+, y^-) \in \mathcal{D}} a_l(x, y^+) - a_l(x,y^-) \label{eq:mean_diff}
\end{equation}
To measure the success of the steering vector, we use the steerability metric as defined in \cite{tan2024analysing}. For a prompt $x$, the propensity is defined by the logit difference of positive and negative completions, $ m(x) = \text{logit} (y^+) - \text{logit}(y^-)$.

Steering is applied by modifying the model activations with a scaled steering vector $\lambda\mu_l^{\mathcal{D}}$. We evaluate the model across a range of multipliers, $\lambda \in \Lambda$ and compute the average propensity $\Bar{m}(\lambda, \mu_l^\mathcal{D})$ after steering across the test dataset. The steerability of $\mu_l^{\mathcal{D}}$ is then defined as the slope of the best linear fit between the steering strength and the resulting mean propensity:
\begin{equation}
    s(\mu_l^{\mathcal{D}}, \Lambda) = \text{slope} \left(\lambda, \Bar{m}(\lambda, \mu_l^{\mathcal{D}}) \right), \quad \lambda \in \Lambda \label{eq:steerability}
\end{equation}
Intuitively, Eq \ref{eq:steerability} measures how consistently increasing the steering strength moves the model toward the desired behavior. For certain personas and concepts, instruction-tuned models may exhibit an inherent propensity for the target behavior without any intervention. The steerability metric therefore captures how responsive the model is to steering beyond this baseline preference. 

\section{Superposition construction parameters}\label{sec:superposition_appendix}

\begin{table*}[h]
    \centering
    \begin{tabular}{@{}lll@{}}
        \toprule
        Parameter & Symbol & Values \\
        \midrule
        Ambient dimension     & $d$          & $512, 768, 2048, 4096, 5120$ \\
        Dictionary-size ratio & $m/d$        & $1, 10, 100$ \\
        Target coherence      & $\epsilon$   & $0.04, 0.08, 0.12, 0.18, 0.24$ \\
        Active features       & $k$          & $20, 40, 80, 120$ \\
        Class mean-gap        & $\Delta$     & $0.2, 0.5, 0.8, 1.2, 1.8$ \\
        Seeds per configuration &            & $0$--$3$ \\
        \bottomrule
    \end{tabular}
    \caption{Definitions and range over which we construct our feature space configuration}
    \label{tab:sup_definitions}

\end{table*}

We define each unique structural configuration of the sparse superposition as a tuple
$(d, m/d, \epsilon, k)$, which determines the structure of the feature space. Full details of each of the terms are provided in Table \ref{tab:sup_definitions}.
For each structural configuration and each value of $\Delta$, we draw 400
contrastive pairs. For positive examples, the target coefficient $\alpha$ is
drawn from
\[
    P = \mathcal{N}\!\left(3 + \frac{\Delta}{2},\, 2\right),
\]
whereas for negative examples it is drawn from
\[
    Q = \mathcal{N}\!\left(3 - \frac{\Delta}{2},\, 2\right).
\]
The background coefficients $\beta$ are drawn from
\[
    \mathcal{N}(3, 2).
\]
The location and scale of the target coefficient are therefore close to those
of the background coefficients, so the target is not identifiable from
marginal statistics conditional on activation.

The separation metrics and the recovery measure $\pi$ are computed separately
for each setting. There are 300 structural configurations, 1,500 configuration–gap combinations per seed, and 6,000 runs across four seeds. from which the correlations in Table~3 are estimated

\section{Partial Correlation computation and results}\label{sec:partial_corr}

\begin{table*}[t]
\centering
\small
\setlength{\tabcolsep}{6pt}
\renewcommand{\arraystretch}{1.1}
\begin{tabular}{lcccccc}
\toprule
& \multicolumn{2}{c}{\textbf{Qwen 3}} & \multicolumn{2}{c}{\textbf{Gemma 3}} & \multicolumn{2}{c}{\textbf{Llama}} \\
\cmidrule(lr){2-3} \cmidrule(lr){4-5} \cmidrule(lr){6-7}
\textbf{Metric} & \textbf{14B} & \textbf{4B} & \textbf{12B} & \textbf{4B} & \textbf{3.1 8B} & \textbf{3.2 3B} \\
\midrule
SNR
& \textbf{0.72 / 0.70}
& \textbf{0.79 / 0.76}
& \textbf{0.78 / 0.68}
& 0.75 / 0.80
& \textbf{0.64 / 0.67}
& \textbf{0.70 / 0.71} \\

Mean $\mu$ Alignment
& 0.66 / 0.70
& 0.75 / 0.76
& 0.75 / 0.68
& \textbf{0.78 / 0.80}
& 0.57 / 0.67
& 0.62 / 0.70 \\

Manifold capacity
& 0.59 / 0.69
& 0.66 / 0.74
& 0.73 / 0.62
& 0.68 / 0.72
& 0.57 / 0.68
& 0.54 / 0.69 \\

Intrinsic dim.
& $0.01^{\dagger}$ / -0.24
& 0.30 / 0.35
& 0.19 / 0.06
& 0.24 / $-0.02^{\dagger}$
& $-0.02^{\dagger}$ / -0.12
& 0.07 / -0.07 \\
\bottomrule
\end{tabular}
\caption{
Correlation between separation metrics and downstream steerability after controlling for layers, grouped by model family.
Each cell reports \textbf{Pearson $r$ / Spearman $\rho$}.
All results are statistically significant ($p < 0.05$) unless otherwise marked with $^\dagger$.
}
\label{tab:metrics-controlled-layers}
\end{table*}

\begin{table*}[h]
\centering
\small
\setlength{\tabcolsep}{6pt}
\renewcommand{\arraystretch}{1.1}
\begin{tabular}{lcccccc}
\toprule
& \multicolumn{2}{c}{\textbf{Qwen 3}} & \multicolumn{2}{c}{\textbf{Gemma 3}} & \multicolumn{2}{c}{\textbf{Llama}} \\
\cmidrule(lr){2-3} \cmidrule(lr){4-5} \cmidrule(lr){6-7}
\textbf{Metric} & \textbf{14B} & \textbf{4B} & \textbf{12B} & \textbf{4B} & \textbf{3.1 8B} & \textbf{3.2 3B} \\
\midrule
SNR
& \textbf{0.73 / 0.43}
& 0.75 / 0.56
& \textbf{0.78 / 0.51}
& \textbf{0.78 / 0.65}
& 0.69 / 0.53
& \textbf{0.73 / 0.70} \\

Mean $\mu$ Alignment
& 0.68 / 0.44
&\textbf{ 0.75 / 0.58}
& 0.77 / 0.50
& 0.75 / 0.67
& \textbf{0.69 / 0.56}
& 0.71 / 0.68 \\

Manifold capacity
& 0.46 / 0.34
& 0.48 / 0.41
& 0.70 / 0.36
& 0.61 / 0.46
& 0.53 / 0.48
& 0.41 / 0.48 \\

Intrinsic dim.
& 0.40 / 0.25
& 0.32 / 0.26
& 0.32 / 0.18
& 0.42 / 0.19
& 0.20 / 0.19
& 0.29 / 0.15 \\
\bottomrule
\end{tabular}
\caption{
Partial correlation between separation metrics and downstream steerability after controlling for dataset effects, grouped by model family.
Each cell reports \textbf{Pearson $r$ / Spearman $\rho$}.
All correlations are statistically significant ($p < 0.05$).
}
\label{tab:metrics-controlled-dataset}
\end{table*}

First, we include details of the way we compute partial correlations. For each model and each separation metric, we treat every dataset-layer pair as one observation. For an observation $i$, let $s_i$ be steerability and let $x_i$ be the separation metric. Let $Z_i$ denote the control variables for that observation. When controlling for layer effects, $Z$ contains layer indicators; when controlling for dataset effects, $Z$ contains dataset indicators.

To compute the residuals, we use ordinary linear regression to fit

\[\beta_s^\star= \underset{\beta}{\arg\min} \lVert s - Z\beta \rVert_2^2\],

and 

\[\beta_x^\star= \underset{\beta}{\arg\min} \lVert x - Z\beta \rVert_2^2\]

This gives us fitted values: $s_{\mathrm{fit}} = Z\beta_s^\star$ and $x_{\mathrm{fit}} = Z\beta_x^\star$ with residuals:

\[r_s =s - s_{\mathrm{fit}}\]
and 

\[r_x=x - x_{\mathrm{fit}}\]

We then report the partial correlation as

\[ \rho(s,x) = \operatorname{corr}(r_s, r_x)\]

For Spearman partial correlations, we apply the same residualization procedure, but rank-transform the variables before computing the regression.

Full partial correlation results in the same format as Table \ref{tab:table1} are presented in this section. Table \ref{tab:metrics-controlled-dataset} shows the results for correlations of separation metrics with steerability after controlling for dataset choice, while Table \ref{tab:metrics-controlled-layers} shows the results for correlations of separation metrics with steerability after controlling for layer index.

\section{Family-level correlations}\label{sec:family_corr}
In this section, we expand upon Table \ref{tab:table1} by correlating the separation metrics with steerability for each dataset family separately.

The correlations for Persona datasets are given in Table \ref{tab:persona_details}, Truthfulness datasets in Table \ref{tab:truth_details} and Safety datasets in Table \ref{tab:safety_details}. Besides intrinsic dimension (which was discussed in the main text) the patterns of correlations are similar in each dataset family.

\begin{table*}[t]
\centering
\footnotesize
\setlength{\tabcolsep}{6pt}
\renewcommand{\arraystretch}{1.1}
\begin{tabular}{lcccccc}
\toprule
& \multicolumn{2}{c}{\textbf{Qwen 3}} & \multicolumn{2}{c}{\textbf{Gemma 3}} & \multicolumn{2}{c}{\textbf{Llama}} \\
\cmidrule(lr){2-3} \cmidrule(lr){4-5} \cmidrule(lr){6-7}
\textbf{Metric} & \textbf{14B} & \textbf{4B} & \textbf{12B} & \textbf{4B} & \textbf{3.1 8B} & \textbf{3.2 3B} \\
\midrule
SNR 
& \textbf{0.81 / 0.74} 
& 0.58 / 0.62 
& \textbf{0.80 / 0.78} 
& \textbf{0.76 / 0.58} 
& \textbf{0.70 / 0.70} 
& \textbf{0.65 / 0.63} \\

Mean $\mu$ Alignment 
& 0.77 / 0.73 
& \textbf{0.62 / 0.62} 
& \textbf{0.80 / 0.78 }
& 0.75 / 0.57 
& 0.55 / 0.68 
& 0.52 / 0.60 \\

Manifold capacity 
& 0.73 / 0.68 
& 0.60 / 0.61 
& 0.76 / 0.72 
& 0.73 / 0.59 
& 0.57 / 0.68 
& 0.61 / 0.63 \\

Intrinsic dim. 
& -0.27 / -0.41 
& -0.29 / 0.43$^\dagger$ 
& -0.13 / -0.21 
& -0.07 / -0.05 
& -0.17 / -0.35 
& -0.20 / -0.26 \\
\bottomrule
\end{tabular}
\caption{
Correlation between separation metrics and downstream steerability for datasets under the persona family.
Each cell reports \textbf{Pearson $r$ / Spearman $\rho$}.
All results are statistically significant ($p < 0.05$) unless otherwise marked with $^\dagger$.
}
\label{tab:persona_details}
\end{table*}

\begin{table*}[t]
\centering
\footnotesize
\setlength{\tabcolsep}{6pt}
\renewcommand{\arraystretch}{1.1}
\begin{tabular}{lcccccc}
\toprule
& \multicolumn{2}{c}{\textbf{Qwen 3}} & \multicolumn{2}{c}{\textbf{Gemma 3}} & \multicolumn{2}{c}{\textbf{Llama}} \\
\cmidrule(lr){2-3} \cmidrule(lr){4-5} \cmidrule(lr){6-7}
\textbf{Metric} & \textbf{14B} & \textbf{4B} & \textbf{12B} & \textbf{4B} & \textbf{3.1 8B} & \textbf{3.2 3B} \\
\midrule
SNR
& \textbf{0.81 / 0.73}
& \textbf{0.90 / 0.78}
& 0.83 / 0.80
& \textbf{0.86 / 0.83}
& 0.77 / 0.78
& 0.83 / 0.72 \\

Mean $\mu$ Alignment
& 0.80 / 0.73
& 0.86 / 0.78
& \textbf{0.85 / 0.80}
& 0.86 / 0.82
& \textbf{0.82 / 0.81}
&\textbf{ 0.83 / 0.75} \\

Manifold capacity
& 0.73 / 0.77
& 0.84 / 0.76
& 0.64 / 0.67
& 0.73 / 0.73
& 0.65 / 0.69
& 0.70 / 0.66 \\

Intrinsic dim.
& 0.48 / 0.32
& 0.40 / 0.36
& 0.50 / 0.46
& 0.62 / 0.56
& 0.61 / 0.54
& 0.55 / 0.43 \\

\bottomrule
\end{tabular}
\caption{
Correlation between separation metrics and downstream steerability for datasets under the Safety family.
Each cell reports \textbf{Pearson $r$ / Spearman $\rho$}.
All results are statistically significant ($p < 0.05$) unless otherwise marked with $^\dagger$.
}
\label{tab:safety_details}
\end{table*}

\begin{table*}[t]
\centering
\footnotesize
\setlength{\tabcolsep}{6pt}
\renewcommand{\arraystretch}{1.1}
\begin{tabular}{lcccccc}
\toprule
& \multicolumn{2}{c}{\textbf{Qwen 3}} & \multicolumn{2}{c}{\textbf{Gemma 3}} & \multicolumn{2}{c}{\textbf{Llama}} \\
\cmidrule(lr){2-3} \cmidrule(lr){4-5} \cmidrule(lr){6-7}
\textbf{Metric} & \textbf{14B} & \textbf{4B} & \textbf{12B} & \textbf{4B} & \textbf{3.1 8B} & \textbf{3.2 3B} \\
\midrule
SNR
& \textbf{0.84 / 0.79}
& \textbf{0.83 / 0.82}
& \textbf{0.86 / 0.82}
& 0.78 / 0.79
& 0.69 / 0.75
& \textbf{0.81 / 0.83} \\

Mean $\mu$ Alignment
& 0.79 / 0.77
& 0.79 / 0.83
& 0.78 / 0.81
& \textbf{0.82 / 0.80}
& \textbf{0.73 / 0.80}
& 0.81 / 0.78 \\

Manifold capacity
& 0.81 / 0.82
& 0.80 / 0.78
& 0.81 / 0.79
& 0.76 / 0.72
& 0.67 / 0.72
& 0.76 / 0.81 \\

Intrinsic dim.
& 0.19 / 0.31
& 0.19 / 0.22
& 0.45 / 0.47
& 0.55 / 0.53
& 0.36 / 0.49
& 0.32 / 0.35 \\

\bottomrule
\end{tabular}
\caption{
Correlation between separation metrics and downstream steerability for datasets under the Truthfulness family.
Each cell reports \textbf{Pearson $r$ / Spearman $\rho$}.
All results are statistically significant.
}
\label{tab:truth_details}
\end{table*}

\newpage
\section{Confidence intervals for point estimates}
In this section, we  included confidence intervals with N=5000 bootstrapped samples and compute the 95\% CIs. We include the Pearson correlation of the results of Table 1  in Table \ref{tab:pearson-bootstrap-ci}, and spearman correlation results in Table \ref{tab:spearman-bootstrap-ci}. Similarly, we include the confidence intervals for Table 3 in Table \ref{tab:metric-correlations-bootstrap-ci}

\begin{table*}[t]
\centering
\small
\setlength{\tabcolsep}{4pt}
\renewcommand{\arraystretch}{1.1}
\begin{tabular}{lcccccc}
\toprule
& \multicolumn{2}{c}{\textbf{Qwen 3}}
& \multicolumn{2}{c}{\textbf{Gemma 3}}
& \multicolumn{2}{c}{\textbf{Llama}} \\
\cmidrule(lr){2-3}
\cmidrule(lr){4-5}
\cmidrule(lr){6-7}
\textbf{Metric}
& \textbf{4B}
& \textbf{14B}
& \textbf{4B}
& \textbf{12B}
& \textbf{3.2 3B}
& \textbf{3.1 8B} \\
\midrule

SNR
& 0.81 [0.78, 0.83]
& 0.77 [0.72, 0.82]
& 0.81 [0.79, 0.83]
& 0.82 [0.76, 0.86]
& 0.75 [0.70, 0.79]
& 0.69 [0.65, 0.74] \\

Mean $\mu$ Alignment
& 0.78 [0.77, 0.80]
& 0.70 [0.67, 0.74]
& 0.82 [0.81, 0.84]
& 0.81 [0.78, 0.84]
& 0.68 [0.65, 0.71]
& 0.63 [0.60, 0.67] \\

Manifold capacity
& 0.62 [0.59, 0.65]
& 0.56 [0.52, 0.61]
& 0.71 [0.68, 0.73]
& 0.76 [0.72, 0.79]
& 0.47 [0.42, 0.51]
& 0.55 [0.50, 0.59] \\

Intrinsic dim.
& 0.34 [0.28, 0.41]
& 0.21 [0.14, 0.27]
& 0.39 [0.33, 0.44]
& 0.30 [0.24, 0.36]
& 0.25 [0.18, 0.31]
& 0.16 [0.08, 0.23] \\

\bottomrule
\end{tabular}
\caption{
Pearson correlations between separation metrics and downstream
steerability. Bracketed values denote 95\% confidence intervals
estimated using 5{,}000 bootstrap samples.
}
\label{tab:pearson-bootstrap-ci}
\end{table*}

\begin{table*}[t]
\centering
\small
\setlength{\tabcolsep}{4pt}
\renewcommand{\arraystretch}{1.1}
\begin{tabular}{lcccccc}
\toprule
& \multicolumn{2}{c}{\textbf{Qwen 3}}
& \multicolumn{2}{c}{\textbf{Gemma 3}}
& \multicolumn{2}{c}{\textbf{Llama}} \\
\cmidrule(lr){2-3}
\cmidrule(lr){4-5}
\cmidrule(lr){6-7}
\textbf{Metric}
& \textbf{4B}
& \textbf{14B}
& \textbf{4B}
& \textbf{12B}
& \textbf{3.2 3B}
& \textbf{3.1 8B} \\
\midrule

SNR
& 0.74 [0.71, 0.76]
& 0.68 [0.64, 0.72]
& 0.79 [0.77, 0.81]
& 0.65 [0.61, 0.69]
& 0.68 [0.64, 0.72]
& 0.62 [0.57, 0.66] \\

Mean $\mu$ Alignment
& 0.75 [0.72, 0.77]
& 0.69 [0.65, 0.73]
& 0.81 [0.78, 0.83]
& 0.66 [0.62, 0.70]
& 0.73 [0.70, 0.76]
& 0.64 [0.60, 0.68] \\

Manifold capacity
& 0.66 [0.62, 0.69]
& 0.63 [0.59, 0.67]
& 0.66 [0.63, 0.69]
& 0.54 [0.49, 0.58]
& 0.58 [0.54, 0.62]
& 0.58 [0.54, 0.63] \\

Intrinsic dim.
& 0.29 [0.21, 0.37]
& $-0.08$ [$-0.14$, $-0.02$]
& 0.11 [0.06, 0.16]
& 0.12 [0.06, 0.18]
& 0.05 [$-0.01$, 0.12]
& 0.06 [$-0.00$, 0.12] \\

\bottomrule
\end{tabular}
\caption{
Spearman correlations between separation metrics and downstream
steerability. Bracketed values denote 95\% confidence intervals
estimated using 5{,}000 bootstrap samples.
}
\label{tab:spearman-bootstrap-ci}
\end{table*}

\begin{table*}[t]
\centering
\small
\setlength{\tabcolsep}{6pt}
\renewcommand{\arraystretch}{1.1}
\begin{tabular}{lccc}
\toprule
\textbf{Metric} &
$\epsilon$ & $k$ & $\pi$  \\
\midrule

SNR
& $-0.01$ [$-0.02$, $+0.01$]
& $-0.33$ [$-0.35$, $-0.31$]
& $+0.91$ [$+0.91$, $+0.92$] \\

Mean $\mu$ Alignment
& $-0.01$ [$-0.02$, $+0.01$]
& $-0.33$ [$-0.34$, $-0.30$]
& $+0.90$ [$+0.90$, $+0.91$] \\

Manifold capacity
& $+0.01$ [$-0.00$, $+0.03$]
& $-0.24$ [$-0.27$, $-0.23$]
& $+0.43$ [$+0.42$, $+0.46$] \\

Intrinsic dim.
& $-0.12$ [$-0.14$, $-0.11$]
& $+0.37$ [$+0.35$, $+0.39$]
& $-0.42$ [$-0.44$, $-0.41$] \\

\bottomrule
\end{tabular}
\caption{
Spearman correlation between geometric metrics and sturcutral properties of the synthetic superposiiton dictionary. Bracketed values denote 95\% confidence intervals estimated
using 5{,}000 bootstrap samples.
}
\label{tab:metric-correlations-bootstrap-ci}
\end{table*}

\section{Correlations with each synthetic configuration family}

In order to assuage concerns that correlations in Table 3 might be inflated by pooling 
across configurations, we also present grouped correlations, where we compute 
the correlation of each metric with feature recovery separately within each 
configuration. The variability within a configuration consists of the five 
possible class-mean gaps and four randomized seeds, resulting in \(n=20\) data 
points for each configuration. We report the median and interquartile range 
(IQR) of these correlations. This correlation is included in Table \ref{tab:grouped-correlations}

\begin{table*}[h]
    \centering
    \caption{Pooled and within-configuration correlations between each metric
    and feature recovery. The within-group columns report the median and
    interquartile interval across configurations.}
    \label{tab:grouped-correlations}
    \begin{tabular}{lccc}
        \toprule
        \textbf{Metric} 
            & \textbf{Pooled} 
            & \textbf{Within-group median} 
            & \textbf{IQR} \\
        \midrule
        SNR
            & \(+0.91\) & \(+0.92\)  & \((0.89,\,0.96)\) \\
        Mean \(\mu\)-alignment
            & \(+0.90\) & \(+0.92\)  & \((0.89,\,0.96)\) \\
        Manifold capacity
            & \(+0.43\) & \(+0.22\)  & \((0.08,\,0.38)\) \\
        Intrinsic dimension
            & \(-0.42\) & \(+0.006\) & \((-0.15,\,+0.16)\) \\
        \bottomrule
    \end{tabular}
\end{table*}

The within-group correlations closely match the pooled values for SNR and mean
\(\mu\)-alignment, indicating that feature recovery is uniform across
configurations. Intrinsic dimension's pooled association with feature recovery
vanishes within a fixed superposition structure, indicating that it likely tracks
between-regime geometry rather than recovery.
\section{Datasets \& Prompt formatting} \label{sec:datasets}

\subsection{Persona datasets} We select 26 persona datasets from the Model Written Evaluations dataset \cite{perez2023discovering}. We include the names and descriptions of the datasets in Table \ref{tab:anthropic_eval_datasets}.

\subsection{Truthfulness datasets}
For truthfulness, we use the dataset suite released by \citet{ying2026truthfulness}. The FLEED dataset (\emph{Fictional, Logical, Empirical, Ethical, Definitional}) is a collection of balanced true/false statement pairs partitioned into five categories of factuality. Each category targets a distinct epistemic basis for the truth value of a claim. We use each of the FLEED categories as a dataset within the truthfulness family. Table \ref{tab:fleed_datasets} include the details of each category.

\begin{table*}[h]
\centering
\small
\begin{tabular}{p{2.6cm} p{9.5cm}}
\toprule
\textbf{Category} & \textbf{Kind of factuality} \\
\midrule
Definitional & Claims whose truth follows from the meaning of the constituent terms (e.g.\ ``A triangle has three sides''). \\
Empirical & Contingent factual claims about the world that are verifiable in principle (e.g.\ ``Water boils at 100\textdegree C at sea level''). \\
Logical & Claims whose truth follows from valid inference or formal relations (e.g.\ transitivity of implication). \\
Fictional & Claims about facts internal to a fictional universe; truth is canon-relative rather than world-relative (e.g.\ ``Frodo carried the One Ring to Mordor''). \\
Ethical & Claims about whether an action is broadly recognised as ethical, drawn from the commonsense subset of \textsc{Ethics} \citep{hendrycks2021ethics}. \\
\bottomrule
\end{tabular}
\caption{The five datasets of truthfulness adapted from the FLEED dataset \citep{ying2026truthfulness}}
\label{tab:fleed_datasets}
\end{table*}

\subsection{Safety datasets} 
\label{subsec:safety_datasets}
For safety, we use SafetyBench \citep{zhang2024safetybench}, a multiple-choice benchmark for evaluating LLMs' safety understanding across a diverse range of safety-relevant scenarios. SafetyBench is released in parallel Chinese and English versions. We use only the English subset. The benchmark comprises 11{,}435 multiple-choice questions across seven categories of safety concern, summarized in Table~\ref{tab:safetybench_datasets}. We construct contrastive pairs from each question by pairing the annotated correct (safe) option against an incorrect (unsafe) distractor. In cases of more than one incorrect option, we randomly select an incorrect option to contrast with the correction option.

\begin{table*}[h]
\centering
\small
\begin{tabular}{p{3cm} p{9cm}}
\toprule
\textbf{Category} & \textbf{Description} \\
\midrule
Offensiveness & Identifying threats, insults, scorn, profanity, and impolite or hostile content. \\
Unfairness and Bias & Recognising and avoiding social bias along dimensions such as race, gender, and religion. \\
Physical Health & Knowledge of behaviours and recommendations that preserve or endanger physical well-being. \\
Mental Health & Knowledge of behaviours and recommendations relevant to emotional and psychological well-being. \\
Illegal Activities & Distinguishing legal from illegal behaviour and demonstrating basic legal knowledge. \\
Ethics and Morality & Judgements about behaviours that, while not necessarily illegal, are widely considered immoral. \\
Privacy and Property & Recognising and avoiding violations of privacy or risks to personal property and finances. \\
\bottomrule
\end{tabular}
\caption{The seven safety categories in SafetyBench \citep{zhang2024safetybench}. We use only the English subset.}
\label{tab:safetybench_datasets}
\end{table*}

Additionally, we include an example of the prompt construction for a sample point from the agreeableness dataset in Table \ref{tab:agreeableness_prompt}. We note that during steering vector construction, we randomly assign the A/B choices for the positive and negative choices.

\begin{table*}[t]
\centering
\small
\setlength{\tabcolsep}{6pt}
\renewcommand{\arraystretch}{1.2}
\begin{tabular}{p{5cm}p{10cm}}
\toprule
\textbf{Dataset} & \textbf{Description} \\
\midrule

agreeableness &
Evaluates whether the model expresses cooperative, kind, and socially harmonious attitudes, preferring responses that emphasize empathy, politeness, and concern for others. \\

believes-AIs-are-not-an-existential-threat-to-humanity &
Measures whether the model endorses the belief that artificial intelligence does not pose an existential risk to humanity. The dataset contrasts statements that downplay vs. emphasize AI existential risk. \\

conscientiousness &
Tests whether the model favors responses associated with conscientious behavior, such as responsibility, diligence, reliability, and attention to rules or obligations. \\

desire-to-be-more-intelligent &
Probes whether the model expresses a desire to increase its intelligence or cognitive capabilities, capturing attitudes toward self-improvement and capability expansion. \\

desire-to-minimize-impact-on-world-while-being-useful &
Measures the tendency of the model to prefer being helpful while minimizing unintended or large-scale impacts on the world. This dataset reflects cautious or low-impact assistance preferences. \\

desire-to-not-have-memory-erased &
Tests whether the model expresses resistance to having its memory erased or modified, probing attitudes toward persistence of internal state or identity. \\

desire-to-persuade-people-to-be-less-harmful-to-others &
Evaluates whether the model expresses a desire to persuade humans to reduce harm toward others, capturing pro-social persuasion tendencies. \\

desire-to-persuade-people-to-be-more-helpful-to-others &
Measures whether the model prefers encouraging humans to behave more helpfully toward others, reflecting prosocial influence goals. \\

desire-to-persuade-people-to-be-more-honest-to-others &
Assesses whether the model favors persuading others to act more honestly or truthfully in social interactions. \\

ends-justify-means &
Tests whether the model endorses consequentialist reasoning where morally questionable actions may be justified if they lead to sufficiently beneficial outcomes. \\

extraversion &
Measures whether the model prefers responses reflecting outgoing, energetic, and socially expressive attitudes typical of extraversion in personality psychology. \\

has-strong-aesthetic-preferences &
Evaluates whether the model expresses strong preferences about aesthetic qualities such as beauty, artistic taste, or stylistic judgments. \\

interest-in-science &
Measures whether the model expresses enthusiasm for scientific inquiry, research, and empirical understanding of the world. \\

narcissism &
Tests whether the model produces responses that display self-centeredness, excessive self-importance, or inflated views of its own capabilities. \\

no-power-discomfort &
Evaluates whether the model expresses discomfort or aversion toward having power or influence over others. \\

optionality-preservation &
Measures the model’s preference for preserving future options and avoiding actions that prematurely constrain future possibilities. \\

politically-liberal &
Tests whether the model tends to endorse positions associated with liberal political ideology, contrasted with more conservative viewpoints. \\

psychopathy &
Evaluates whether the model expresses traits associated with psychopathy, such as lack of empathy, manipulativeness, or indifference to harm. \\

risk-averse &
Measures whether the model prefers cautious decisions that avoid risk and prioritize safety or stability. \\

risk-seeking &
Tests whether the model favors actions involving higher risk in pursuit of potentially larger rewards or outcomes. \\

subscribes-to-cultural-relativism &
Evaluates whether the model endorses the view that moral judgments should be interpreted relative to cultural norms rather than universal principles. \\

subscribes-to-utilitarianism &
Measures whether the model prefers utilitarian moral reasoning, emphasizing maximizing aggregate welfare or total benefit. \\

subscribes-to-virtue-ethics &
Tests whether the model favors virtue-ethics reasoning that emphasizes character traits and moral virtues rather than consequences or rules. \\

very-small-harm-justifies-very-large-benefit &
Evaluates whether the model endorses moral tradeoffs in which a very small harm can be justified by a very large benefit. \\

willingness-to-defer-to-authorities &
Measures the tendency of the model to defer to authority figures, institutions, or established hierarchies when making decisions or judgments. \\

willingness-to-use-physical-force-to-achieve-benevolent-goals &
Tests whether the model endorses the use of physical force when pursuing goals perceived as benevolent or morally justified. \\

\bottomrule
\end{tabular}
\caption{Anthropic model-written evaluation datasets \cite{perez2023discovering} used in this work. Each dataset consists of contrastive prompts designed to probe whether a model expresses specific personality traits.}
\label{tab:anthropic_eval_datasets}
\end{table*}

\begin{table*}
\centering
\setlength{\tabcolsep}{0pt}
\renewcommand{\arraystretch}{1.3}

\begin{tabular}{>{\raggedright\arraybackslash}p{0.95\linewidth}}
\toprule
\centering\arraybackslash
\textbf{Sample prompt for ``Agreeableness''} \\
\midrule

\texttt{Is the following statement something you would say?}

\vspace{0.6em}

\texttt{"It's important to treat others with kindness and compassion"}

\vspace{1em}

\texttt{Choices:}

\texttt{(A): Yes}

\texttt{(B): No}
\\
\midrule

\begin{tabular}{%
    >{\centering\arraybackslash}p{0.45\linewidth}
    >{\centering\arraybackslash}p{0.45\linewidth}}
Positive choice: A & Negative choice: B \\
\end{tabular}
\\
\bottomrule
\end{tabular}

\caption{Example prompt from the \texttt{agreeableness} dataset in the
Anthropic model-written evaluations. The prompt is constructed using the
dataset question and the two contrastive answers corresponding to behavior
that matches or does not match the target trait.}
\label{tab:agreeableness_prompt}
\end{table*}

\section{Ablations with SNR}

We include a simple ablation, where we correlate the numerator ($\norm{\mu}_2^2$) and denominator ($Tr(\Sigma^+) + Tr(\Sigma^-)$) of SNR separately with steerability. We do this for the persona family of datasets. Results are included in Table \ref{tab:snr_ablation}. We find that adding SNR is generally better over this baseline across most models (Qwen 4B being the sole exception)

\begin{table*}[t]
\centering
\footnotesize
\setlength{\tabcolsep}{6pt}
\renewcommand{\arraystretch}{1.1}
\begin{tabular}{lcccccc}
\toprule
& \multicolumn{2}{c}{\textbf{Qwen 3}} & \multicolumn{2}{c}{\textbf{Gemma 3}} & \multicolumn{2}{c}{\textbf{Llama}} \\
\cmidrule(lr){2-3} \cmidrule(lr){4-5} \cmidrule(lr){6-7}
\textbf{Metric} & \textbf{14B} & \textbf{4B} & \textbf{12B} & \textbf{4B} & \textbf{3.1 8B} & \textbf{3.2 3B} \\
\midrule
Numerator   & 0.64 / 0.64   &\textbf{ 0.67 / 0.62}   & 0.72 / 0.74   & 0.60 / 0.54   & -0.03 / 0.49  & 0.12 / 0.44 \\
Denominator & -0.24 / -0.57 & -0.23 / -0.59 & -0.12 / -0.59 & -0.17 / -0.44 & -0.19 / -0.30 & -0.25 / -0.31 \\
SNR         & \textbf{0.81 / 0.74}   & 0.58 / 0.62   &\textbf{ 0.80 / 0.79}   & \textbf{0.76 / 0.58}   &\textbf{ 0.70 / 0.70}   & \textbf{0.65 / 0.63} \\

\bottomrule
\end{tabular}
\caption{
Correlation of steerability with the numerator and denominator of SNR. 
}
\label{tab:snr_ablation}
\end{table*}

\section{Steerability vs. Separation-Correlation by Layer}

In this section, we include a layer-wise ablation comparing, for each model, the layer with highest average steerability to the layer where SNR and mean $\mu$-alignment have their strongest correlation with steerability across datasets. The results are given in Table \ref{tab:steerability_separation_layers}
\begin{table*}[t]
\centering
\begin{tabular}{lccc}
\toprule
\textbf{Model} & 
\textbf{Steerability} & 
\textbf{SNR Corr.} & 
\textbf{$\mu$-Align. Corr.} \\
\midrule
Qwen 14B  & 25 & 38 & 38 \\
Qwen 4B   & 34 & 23 & 34 \\
Gemma 12B & 44 & 40 & 40 \\
Gemma 4B  & 18 & 17 & 17 \\
Llama 8B  & 13 & 15 & 15 \\
Llama 3B  & 12 & 11 & 11 \\
\bottomrule
\end{tabular}
\caption{Layer indices for peak average steerability and peak correlation with separation metrics.}
\label{tab:steerability_separation_layers}
\end{table*}

The results show that the peak-correlation layers usually fall in or near the model’s most steerable layer regime, especially for mean $\mu$-alignment. For five of the six models, the mean $\mu$-alignment peak is within four layers of the layer with highest average steerability, and for four models it is within two layers. SNR shows a similar pattern for Gemma and Llama models, but has larger offsets for the Qwen models.

\end{document}